\documentclass[sigconf]{acmart}

\setcopyright{cc}
\setcctype{by}
\copyrightyear{2026}
\acmYear{2026}
\acmConference[MM '26]{Proceedings of the 34th ACM International Conference on Multimedia}{November 10--14, 2026}{Rio de Janeiro, Brazil}
\acmBooktitle{Proceedings of the 34th ACM International Conference on Multimedia (MM '26), November 10--14, 2026, Rio de Janeiro, Brazil}
\acmISBN{979-8-4007-2213-4/2026/11}
\acmDOI{10.1145/3767308.3836432}
\AtBeginDocument{%
  }

\usepackage{booktabs}
\usepackage{multirow}
\usepackage{amsmath}

\usepackage{amssymb}
\usepackage{siunitx}
\usepackage{graphicx}
\usepackage{subcaption}
\usepackage{enumitem}
\usepackage{array}
\usepackage{url}
\usepackage{xcolor}
\usepackage{placeins}

\newcommand{\reldiff}[1]{{\color{green!50!black}\scriptsize +#1\%}}

\title[WaveLLM]{WaveLLM: Wavelet-Domain LLM Entropy Modeling for Lossless Image Compression}

\author{Yue Yu}
\affiliation{%
  \institution{University of Science and Technology of China}
  \city{Hefei}
  \country{China}
}
\email{yuyue02@mail.ustc.edu.cn}

\author{Bowei Kang}
\affiliation{%
  \department{MOE Key Laboratory of Brain-Inspired Intelligent Perception and Cognition}
  \institution{University of Science and Technology of China}
  \city{Hefei}
  \country{China}
}
\email{kbw2000@mail.ustc.edu.cn}

\author{Changsheng Gao}
\correspondingauthor
\affiliation{%
  \institution{Nanyang Technological University}
  \city{Singapore}
  \country{Singapore}
}
\email{changsheng.gao@ntu.edu.sg}

\author{Dong Liu}
\affiliation{%
  \department{MOE Key Laboratory of Brain-Inspired Intelligent Perception and Cognition}
  \institution{University of Science and Technology of China}
  \city{Hefei}
  \country{China}
}
\email{dongeliu@ustc.edu.cn}

\author{Xiaoyan Sun}
\correspondingauthor
\affiliation{%
  \department{National Engineering Laboratory for Brain-inspired Intelligence Technology and Application;
  MoE Key Laboratory of Brain-inspired Intelligent Perception and Cognition}
  \institution{University of Science and Technology of China}
  \city{Hefei}
  \country{China}
}
\additionalaffiliation{%
  \department{Anhui Province Key Laboratory of Biomedical Imaging and Intelligent Processing}
  \institution{Institute of Artificial Intelligence, Hefei Comprehensive National Science Center}
  \city{Hefei}
  \country{China}
}
\email{sunxiaoyan@ustc.edu.cn}

\renewcommand{\shortauthors}{Yue Yu, Bowei Kang, Changsheng Gao, Dong Liu, \& Xiaoyan Sun}

\begin{document}

\begin{abstract}
Lossless image compression fundamentally relies on accurate entropy modeling, driving a recent shift toward large language model (LLM)-based methods.
However, existing approaches mainly emphasize the modeling capacity of LLMs while directly serializing images into pixel or residual sequences, leaving heterogeneous signal patterns entangled within a single prediction stream. In this paper, we argue that effective entropy modeling depends not only on model capacity, but also on the predictability of the input representation.
Motivated by this insight, we propose WaveLLM, a wavelet-domain LLM-based entropy model for lossless image compression. By decomposing an image into multi-level wavelet subbands, WaveLLM converts a heterogeneous prediction problem into a set of more structured and statistically homogeneous subproblems that are more amenable to autoregressive probability estimation. 
To bridge numerical coding symbols with the LLM's native text token space, we introduce a dedicated numerical interface featuring value embeddings and a coefficient prediction head. Furthermore, a task-adaptive prompting strategy is designed to enable a single shared model to adapt to diverse subbands within a unified framework.
Built on Llama-3.2-3B with LoRA adaptation, WaveLLM achieves 2.51, 2.24, and 2.27 bpsp on Kodak, CLIC.pro, and DIV2K, respectively, outperforming the PixelCNN-based codec using the same wavelet transform by 0.43, 0.36, and 0.46 bpsp, and surpassing recent LLM-based lossless compressors on all three benchmarks.
\end{abstract}



\maketitle

\begin{figure*}[t]
  \centering
  \includegraphics[width=0.97\textwidth]{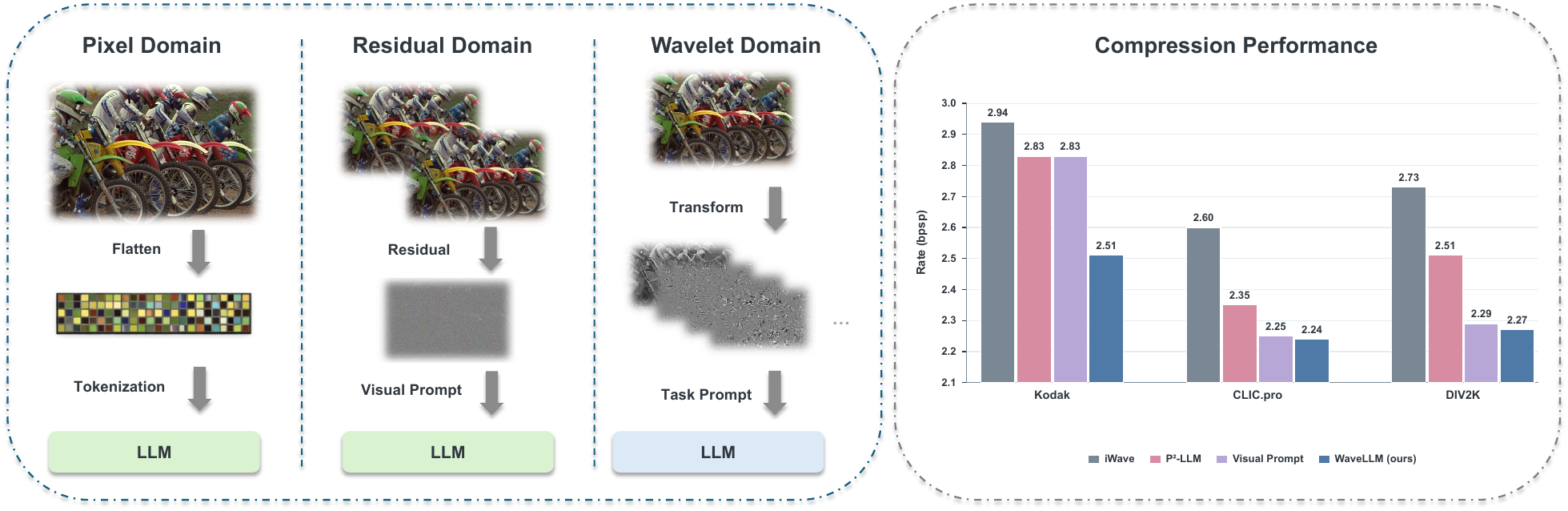}
  \caption{Three paradigms for LLM-based lossless image compression and their performance.
  Left: pixel-domain methods~\cite{chen2024p2llm} linearize the image into a pixel sequence, residual-domain methods~\cite{du2025visualprompt} code residuals from a lossy reconstruction, and WaveLLM decomposes the image into multiresolution wavelet subbands and models their coefficients.
  Right: lossless coding rates (bpsp, lower is better) on three benchmarks.}
  \label{fig_comparison}
  \Description{A teaser figure contrasting three LLM-based compression paradigms and a grouped bar chart comparing coding rates across methods and benchmarks.}
\end{figure*}

\section{Introduction}
Unlike lossy compression, which can trade distortion for bitrate, lossless image compression fundamentally relies on entropy coding. Classical codecs improve this process through carefully designed predictors, reversible transforms, and context models~\cite{png2003,jpegls1998,weinberger2000locoi,sneyers2016flif,taubman2002jpeg2000,jpegxl2019}, while learned methods have further advanced coding efficiency by using neural networks to capture more complex dependencies~\cite{mentzer2019l3c,hoogeboom2019idf,townsend2019hilloc,bai2022dlpr,zhang2021ivpf,zhang2021iflow,zhang2024bps,li2025callic}.
Related work spans image and video coding, feature and representation compression, and other structured signals, with associated datasets, benchmarks, and quality assessments~\cite{huang2021visual,li2024object,sheng2024spatial,huang2026discovering,gao2025dtufc,gao2025cross,sheng2025drfc,li2025ustc,gao2025feature,gao2025cfqa}. These directions provide broader context for our setting, while recent lossless image coding explores large language models (LLMs) as entropy models~\cite{deletang2023lmcompression,chen2024p2llm,du2025visualprompt}.

In this paper, we argue that the accuracy of probability modeling is determined not only by the modeling capacity of the entropy model itself, but also by the predictability of its input. In other words, better entropy coding should not be attributed solely to a stronger model. It also requires an input representation that is more structured, more homogeneous, and thus easier to predict. When heterogeneous signal patterns are entangled within a single sequence, the entropy model must continuously adapt to sharply different statistical regimes, which makes accurate probability estimation intrinsically more difficult. By contrast, when the input is reorganized into simpler components with clearer statistical regularity, the same model can solve a substantially easier prediction problem.

Based on this insight, we propose to perform probability prediction and entropy coding in the wavelet domain. As the work most closely related to ours, we compare three LLM-based methods in Fig. \ref{fig_comparison}. The pixel-domain method directly models pixels~\cite{chen2024p2llm}, requiring the entropy model to handle pixel distributions from different frequency components. The residual-domain method models residuals instead~\cite{du2025visualprompt}, which reduces the difficulty of probability modeling to some extent. In our method, we first transform the image into the wavelet domain and decompose it into progressively organized frequency subbands~\cite{mallat1989,waveletlifting1998,taubman2002jpeg2000}. This decomposition separates coarse low-frequency structures from directional high-frequency details across multiple resolutions~\cite{mallat1989,taubman2002jpeg2000}, thereby converting a highly heterogeneous prediction problem into a collection of more structured and statistically homogeneous subproblems. 

Nevertheless, combining the wavelet domain with an LLM is far from natural. Pretrained LLMs are designed for discrete text tokens, whereas wavelet coefficients are numerical symbols from visual signals. To bridge this gap, we design a dedicated interface that makes wavelet-domain entropy coding compatible with the LLM backbone. Specifically, we introduce a coding-oriented numerical adaptation mechanism that maps wavelet coefficients into a compact local symbol space, equips them with trainable value embeddings, and predicts valid coefficient values through a dedicated value prediction head. This design preserves numerical validity while enabling the pretrained LLM to operate effectively on wavelet-domain symbols.

Another challenge arises from the fact that different wavelet subbands still exhibit distinct statistical characteristics~\cite{wainwright2000scale,taubman2002jpeg2000}. Training an independent entropy model for each subband would significantly increase model complexity and undermine the practicality of LLM-based coding. To address this issue, we further propose a prompt-based adaptive entropy coding strategy, where one shared entropy model is reused across different subbands through task prompts. By conditioning the shared backbone on subband identity and decoded context, the same model adapts its probability predictions to different subband distributions without requiring separate models. In this way, the proposed framework achieves both subband-specific modeling flexibility and parameter efficiency.

Building on the above designs, we present WaveLLM, a wavelet-domain LLM-based entropy model for lossless image compression. The contributions of this work are threefold. 
\begin{itemize}[leftmargin=*,itemsep=2pt,topsep=2pt]
  \item We note the importance of input predictability in lossless image compression and decompose each image into multiple wavelet subbands. The resulting subbands have distinct coefficient statistics that can be modeled separately.
  \item We design a dedicated interface between wavelet-domain signals and an LLM pretrained on language data. It maps coefficients to compact local IDs, uses a dedicated value embedding module, and predicts valid coefficient values over the local range through a dedicated value head.
  \item We propose a task-prompt mechanism for adaptive modeling of different subbands. By conditioning the shared backbone on subband identity and decoded context, WaveLLM adapts to distinct subband statistics without requiring separate models.
\end{itemize}
As shown in Fig. \ref{fig_comparison}, WaveLLM achieves 2.51, 2.24, and 2.27 bpsp on Kodak, CLIC.pro, and DIV2K, respectively. Despite using only a 3B backbone, it outperforms recent 8B LLM-based compressors on all three benchmarks while remaining competitive with strong learned lossless codecs.

\section{Related Work}

\subsection{Lossless Image Compression}
Classical lossless codecs rely on predictive filtering and entropy coding.
PNG~\cite{png2003}, JPEG-LS~\cite{jpegls1998,weinberger2000locoi}, FLIF~\cite{sneyers2016flif}, JPEG2000~\cite{taubman2002jpeg2000}, and JPEG~XL~\cite{jpegxl2019} represent widely used designs. The reversible 5/3 lifting transform in JPEG2000 provides a practical route to lossless wavelet coding~\cite{taubman2002jpeg2000,waveletlifting1998}.
Learned compression methods have explored causal autoregressive models such as PixelCNN and PixelCNN++~\cite{vandenoord2016pixelcnn,salimans2017pixelcnnpp}, hierarchical latent-variable models~\cite{mentzer2019l3c,hoogeboom2019idf,townsend2019hilloc}, invertible flows~\cite{zhang2021ivpf,zhang2021iflow}, residual coding~\cite{mentzer2020learning,bai2022dlpr}, and bit-plane decompositions~\cite{zhang2024bps}.
Across these approaches, the central challenge remains accurate probability estimation over high-dimensional discrete image signals. CALLIC~\cite{li2025callic} reports strong learned-method results through content-adaptive per-image optimization.

\subsection{Wavelet-Domain Codecs}
Wavelets decompose an image into low-frequency structure and high-frequency detail at multiple resolutions~\cite{mallat1989,waveletlifting1998}.
Natural-image wavelet coefficients are non-Gaussian and subband dependent~\cite{wainwright2000scale,taubman2002jpeg2000}, with broader low-frequency responses and sparse, sharply peaked high-frequency coefficients.
This separation makes the wavelet domain attractive for entropy modeling. WaveFormer and BVINet incorporate wavelet modules into Transformer-based video inpainting~\cite{wu2024waveformer,wu2025bvinet}. Unlike these restoration models, WaveLLM uses a reversible wavelet decomposition and decoded same-level subbands for lossless entropy coding.
In lossless coding, JPEG2000 established the canonical framework by combining the reversible 5/3 integer lifting transform with wavelet-domain context modeling and arithmetic coding~\cite{taubman2002jpeg2000}.
The lossless branches of iWave and iWave++ use reversible 5/3 wavelet coding with learned PixelCNN entropy models~\cite{ma2020iwave,ma2022iwavepp}.
WaveLLM studies this setting and replaces the PixelCNN-style entropy model with a shared prompted LLM over codec-defined subband tasks.

\subsection{LLMs for Lossless Image Compression}
Beyond compressing intermediate representations produced by large models \cite{gao2025feature,gao2025dtufc,gao2025cfqa,gao2025cross}, recent LLM-based image compression directly employs pretrained language models as entropy models \cite{deletang2023lmcompression,chen2024p2llm,du2025visualprompt}.
Del\'etang et al.\ demonstrated that autoregressive language modeling can be interpreted directly as compression~\cite{deletang2023lmcompression}.
In image coding, two recent methods apply pretrained LLMs to lossless compression.
P$^2$-LLM keeps the problem in the pixel domain and frames compression as next-pixel prediction~\cite{chen2024p2llm}.
The visual-prompt method introduces a lossy reconstruction as an auxiliary cue and codes the residual with an LLM~\cite{du2025visualprompt}.
Both papers report Llama-3-8B backbones.
WaveLLM takes a different route.
It introduces LLM-based entropy modeling into a reversible 5/3 wavelet codec, using a 3B backbone adapted through LoRA~\cite{hu2022lora}.
Unlike P$^2$-LLM, which processes a rasterized pixel stream without codec-defined structure, and the visual-prompt method, which requires an auxiliary lossy codec to generate visual embeddings, WaveLLM operates within the causal structure of the wavelet codec.
It relies on decoded context that is already available at each coding stage.
At the coefficient interface, it also differs from the reserved-vocabulary strategy used by P$^2$-LLM by introducing a dedicated local value-ID space together with separate value embedding and prediction modules.

\section{Method}

\subsection{Overall Framework}
\label{sec:overview}

\begin{figure*}[t]
  \centering
  \includegraphics[width=0.97\textwidth]{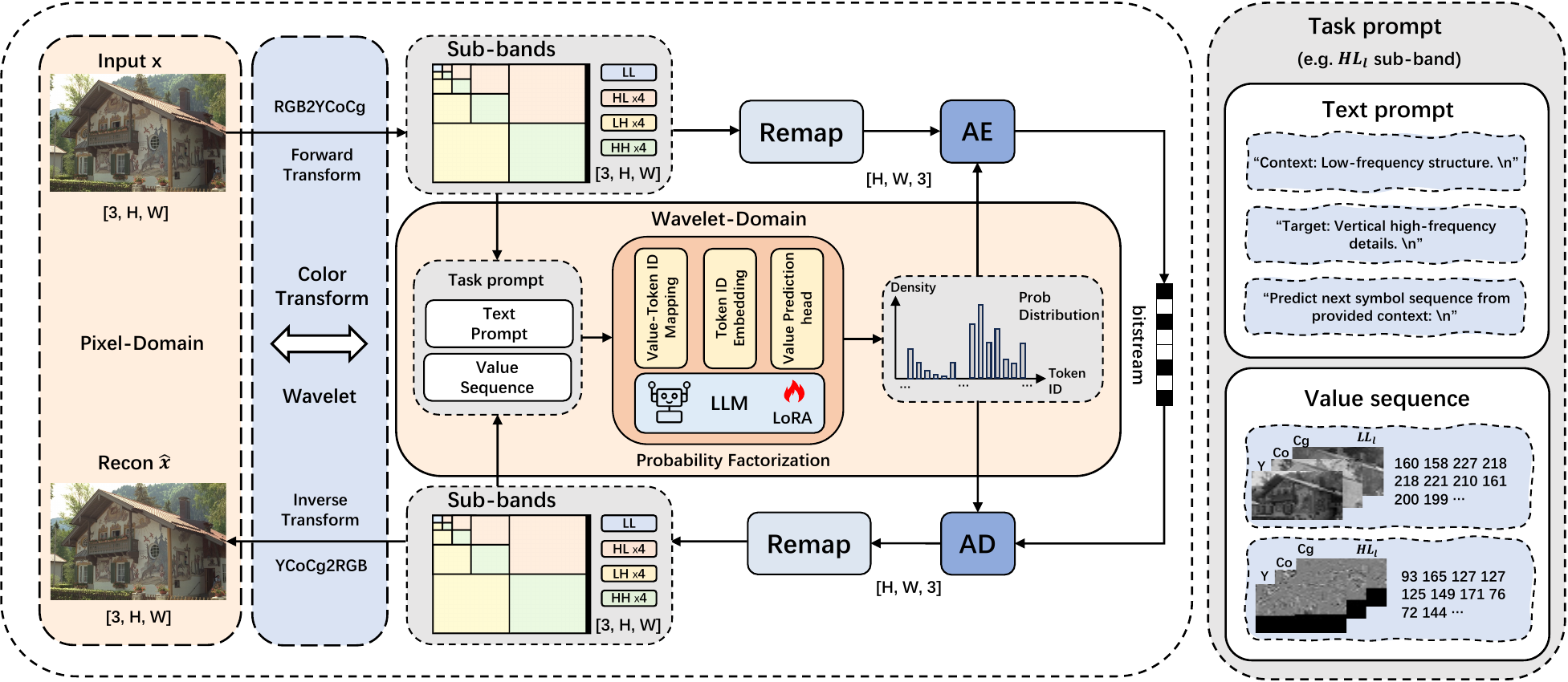}
  \caption{WaveLLM framework.
  Left: WaveLLM converts the input image to YCoCg, performs wavelet decomposition, and entropy-codes the resulting subbands using arithmetic coding.
  Within this coding pipeline, the learnable entropy-modeling core consists of task prompt construction, the coding-oriented numerical interface, and a shared LoRA-adapted LLM that outputs coefficient probability distributions.
  Decoding follows the symmetric path for exact reconstruction.
  Right: expanded task prompt for a representative $HL_l$ task, including the natural-language text prompt and the serialized value sequence formed by same-level $LL_l$ context and $HL_l$ target values.}
  \label{fig:pipeline}
  \Description{A framework diagram of WaveLLM. Left: the WaveLLM coding pipeline with color transform, wavelet decomposition, remapping, arithmetic coding, and exact reconstruction together with the entropy-modeling core. Right: an expanded task prompt for a representative HL subband, showing the text prompt and the serialized value sequence derived from same-level LL context and HL target values.}
\end{figure*}

WaveLLM uses a reversible color transform, a fixed 5/3 wavelet transform, a causal coding order, and arithmetic coding, while the learnable part focuses on wavelet-domain probability factorization over codec-defined subbands.
Figure~\ref{fig:pipeline} summarizes this workflow and highlights the entropy-modeling core composed of task prompts, the coding-oriented numerical interface, and the LLM probability model.

Concretely, an 8-bit RGB image is first converted to YCoCg and decomposed by a four-level fixed 5/3 integer wavelet transform.
This produces 13 subbands: the deepest low-frequency band $LL_4$ and three directional high-frequency bands ($HL_l$, $LH_l$, $HH_l$) at each of four levels.
The reversible 5/3 lifting transform maps integers to integers for lossless coding. Table~\ref{tab:fixed_codec} fixes the transform, coding order, and arithmetic coder while varying only the entropy model.
These subbands are coded in a fixed causal order.
$LL_4$ is coded first, followed by the three high-frequency bands at each level from the deepest level to the shallowest.
For each subband, WaveLLM forms a task prompt, maps the resulting value sequence through the coding-oriented numerical interface, and uses the LLM to produce a probability distribution over coefficient values for arithmetic coding.
The decoder entropy-decodes these subbands in the same order as the encoder. After each level is decoded, the inverse wavelet transform synthesizes the next shallower low-frequency band, ultimately reconstructing the original image exactly.

Before entering the LLM, integer coefficient values are remapped to local value IDs to bridge the gap between wavelet-coefficient arithmetic and the language model's embedding space.
For a three-channel subband tensor $u=[Y,Co,Cg]\in\mathbb{Z}^{3\times h\times w}$, the serialization uses raster order with channel interleaving:
\begin{equation}
s(u)=\left[Y_{1,1},Co_{1,1},Cg_{1,1},Y_{1,2},Co_{1,2},Cg_{1,2},\ldots,Y_{h,w},Co_{h,w},Cg_{h,w}\right].
\end{equation}
The same rule is applied to both context and target bands, ensuring that the autoregressive model always scans spatial positions in a consistent Y/Co/Cg order.
Within the WaveLLM framework, the learnable part highlighted in Figure~\ref{fig:pipeline} is organized around the same three components emphasized in the Introduction.
The following subsections detail these three components: Wavelet-Domain Probability Factorization (Section~\ref{sec:factorization}), Task Prompt Design (Section~\ref{sec:prompt}), and Value Interface and Model Training (Section~\ref{sec:llm}).

\subsection{Wavelet-Domain Probability Factorization}
\label{sec:factorization}

Let $x \in \mathbb{Z}^{3\times H\times W}$ be an 8-bit RGB image.
WaveLLM applies a reversible color transform~$\mathcal{C}$ from RGB to YCoCg~\cite{malvar2008ycocg} and a four-level fixed 5/3 lifting wavelet transform~$\mathcal{W}$~\cite{taubman2002jpeg2000}:
\begin{equation}
\{LL_4,\{HL_l,LH_l,HH_l\}_{l=1}^{4}\}=\mathcal{W}(\mathcal{C}(x)).
\end{equation}
Both transforms are lossless.

This decomposition restructures the entropy-modeling problem in a way favorable to autoregressive modeling.
High-frequency subbands are dominated by near-zero coefficients with sharply peaked distributions, while the low-frequency $LL_4$ band carries the bulk of image energy with broader, more structured statistics.
Separating these two regimes into distinct prediction tasks with explicit causal context removes the need for a single model to simultaneously track coarse structure and fine detail.
This challenge is inherent in flat pixel streams.
Section~\ref{sec:domain_comparison} provides direct experimental evidence for this advantage.

\begin{figure}[t]
  \centering
  \includegraphics[width=0.82\linewidth]{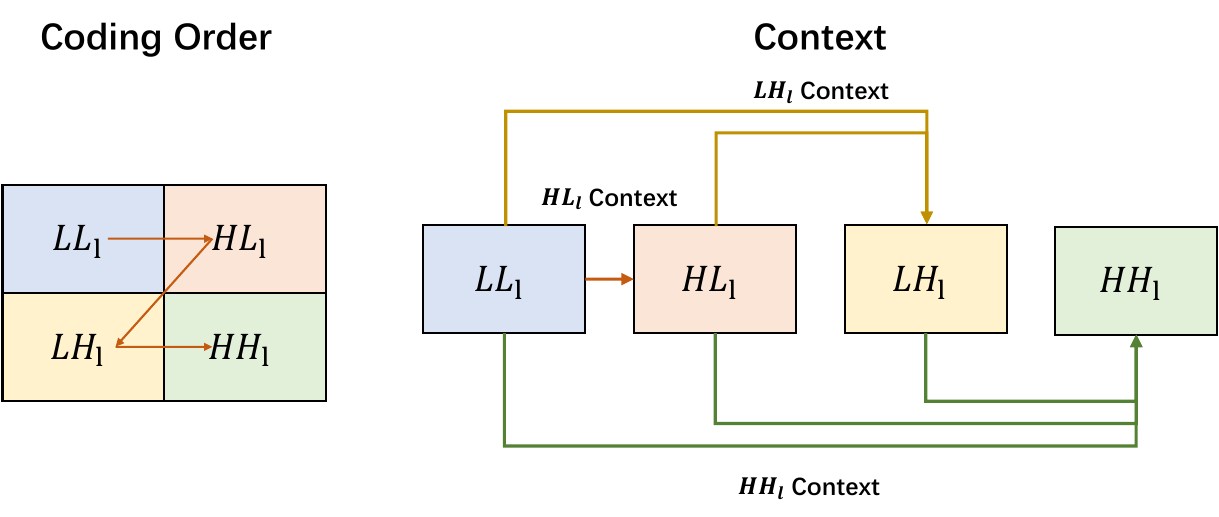}
  \caption{Within-level coding order and same-level context at level $l$. The red arrows indicate $LL_l \rightarrow HL_l \rightarrow LH_l \rightarrow HH_l$; correspondingly, $HL_l$ uses $LL_l$, $LH_l$ uses $[LL_l,HL_l]$, and $HH_l$ uses $[LL_l,HL_l,LH_l]$.}
  \label{fig:dependency}
  \Description{A two-panel diagram showing, at one wavelet level, the spatial arrangement of LL, HL, LH, and HH with red arrows indicating the coding order on the left, and the equivalent same-level dependency graph for the three high-frequency subbands on the right.}
\end{figure}

The coding order follows the lossless wavelet factorization.
$LL_4$ is coded first.
Then, from the deepest level to the shallowest, $HL_l$, $LH_l$, and $HH_l$ are coded in sequence, and the inverse wavelet step at each level reconstructs the next finer-resolution $LL_{l-1}$ band for use as decoded context.
In detail, the codec first models $LL_4$ and then, at each level, predicts $HL_l$, $LH_l$, and $HH_l$ from the already decoded same-level context bands.
Using $C_l^b$ to denote the same-level context for band~$b$ at level~$l$, Figure~\ref{fig:dependency} illustrates this within-level ordering and dependency pattern.
The left panel shows the spatial arrangement of the four subbands, and the red arrows indicate the coding order $LL_l \rightarrow HL_l \rightarrow LH_l \rightarrow HH_l$.
The right panel presents the same structure as a dependency graph:
$C_l^{HL}=LL_l$, $C_l^{LH}=[LL_l,HL_l]$, and $C_l^{HH}=[LL_l,HL_l,LH_l]$.
Through this design, the entropy model remains causal.

The deepest $LL_4$ band receives special treatment.
The MED predictor from JPEG-LS~\cite{weinberger2000locoi} is applied first, and the LLM models the prediction residuals rather than raw coefficient values.
The MED predictor concentrates the $LL_4$ distribution near zero, making the subsequent autoregressive prediction easier.
For high-frequency bands, integer wavelet coefficients are modeled directly.

\subsection{Task Prompt Design}
\label{sec:prompt}

The LLM processes each subband as a one-dimensional sequence.
Each context or target band uses the same raster-order, channel-interleaved serialization defined in Section~\ref{sec:overview}.
Let $y_l^b$ denote the serialized target sequence for subband $b \in \{HL,LH,HH\}$ at level~$l$, and let $r_{LL}$ denote the serialized MED residuals for $LL_4$.

As illustrated in the right panel of Figure~\ref{fig:pipeline}, the LLM input for each subband is organized as a \emph{task prompt} with two components.
Rather than reusing one generic instruction across all coding steps, WaveLLM assigns each subband a task-specific prompt.
The \emph{text prompt} is a task-specific natural-language string that identifies the subband type, describes the available context, and states the prediction objective (e.g., ``Context: Low-frequency structure.\ Target: Vertical high-frequency details.\ Predict:\texttt{\textbackslash n}'').
These tokens are produced by the native Llama tokenizer and processed through the frozen text embedding module.
The text prompt serves not only as a label but as the primary mechanism through which the shared backbone activates different statistical priors for different subbands.
Ablation results in Section~\ref{sec:ablation} show that removing prompts causes the largest single-component degradation.
The \emph{value sequence} contains the serialized integer values of the decoded context and target bands, arranged in causal order and processed through the dedicated value embedding module described in Section~\ref{sec:llm}.
In the representative $HL_l$ example shown in Figure~\ref{fig:pipeline}, same-level $LL_l$ serves as context and $HL_l$ is the target band.
Tasks with multiple context bands follow the same template with additional serialized context segments separated by delimiter tokens.
For $LL_4$, only the text prompt and target MED residuals are needed.

\subsection{Value Interface and Model Training}
\label{sec:llm}

\subsubsection{Coding-Oriented Numerical Interface}

Integer wavelet coefficients should not be represented as ordinary text tokens for two reasons.
First, although the native Llama vocabulary is indexed by consecutive token IDs, those IDs do not encode numerical locality.
Neighboring token IDs correspond to neighboring text symbols rather than neighboring coefficient values.
Second, raw coefficient values include negative integers and therefore cannot be used directly as valid embedding indices or classification labels.
To bridge this gap, each integer symbol~$v$ is mapped to a local value ID
\begin{equation}
z = v + o,
\label{eq:local_id}
\end{equation}
where $z$ denotes the mapped local value ID, and $o=512$ is a fixed offset that shifts the observed coefficient range into the valid local range $\{0,\ldots,K-1\}$ with $K=2048$ while preserving numerical ordering.
Near-zero coefficients, which are the most common values in high-frequency subbands, therefore fall near the center of this local space.
Statistics show that valid coefficients satisfy $v \in [-512, 1535]$.

\begin{figure}[t]
  \centering
  \includegraphics[width=0.95\linewidth]{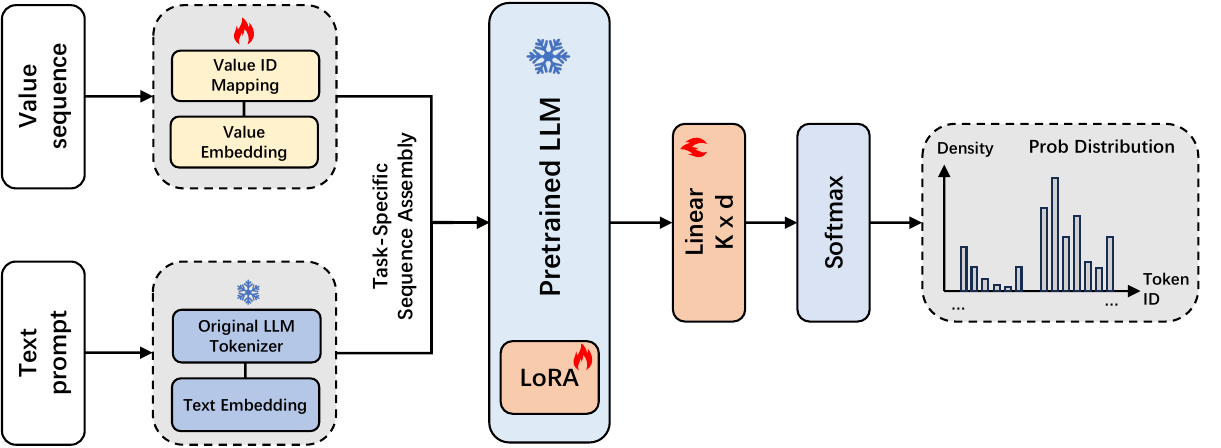}
  \caption{LLM entropy model. Text prompts use the frozen text embedding module, while coefficients are remapped to local IDs and embedded by a dedicated value embedding module. The assembled task sequence is processed by the LoRA-adapted backbone, and a value prediction head outputs probabilities over the local value range for arithmetic coding.}
  \label{fig:value_interface}
  \Description{Architecture diagram showing two input paths (value sequence and text prompt), task-specific sequence assembly, a LoRA-adapted pretrained LLM, and a value prediction head producing a probability distribution.}
\end{figure}

Text prompt tokens and the delimiter token \textit{sep} pass through the frozen text embedding module~$E_{\text{text}}$ (Figure~\ref{fig:value_interface}, lower path).
Coefficient and residual values pass through a trainable value embedding module~$E_{\text{val}} \in \mathbb{R}^{K\times d}$ (upper path), where $d$ is the LLM hidden dimension.
This separation ensures that text semantics and coefficient numerics are processed through independent embedding pathways, avoiding the vocabulary mismatch that arises when coefficient values are treated as ordinary text tokens.
It also differs from the reserved-vocabulary design in P$^2$-LLM. WaveLLM defines a dedicated local value space with its own value embedding and value prediction head.
The two embedding streams are then merged in a task-specific sequence assembly step:
\begin{equation}
X=
\begin{cases}
[E_{\text{text}}(p_{LL}),\, E_{\text{val}}(z(r_{LL}))], & \text{$LL_4$ task},\\[3pt]
[E_{\text{text}}(p_b),\, E_{\text{val}}(z(C_l^b)),\, E_{\text{text}}(\textit{sep}),\, E_{\text{val}}(z(y_l^b))], & \text{HF task},
\end{cases}
\label{eq:input_emb}
\end{equation}
where $p_b$ denotes the text prompt for subband~$b$, $y_l^b$ denotes the serialized target sequence for subband~$b$ at level~$l$, $C_l^b$ denotes the decoded same-level context sequence, and \textit{sep} is a delimiter token between context and target segments.
When $C_l^b$ contains multiple context bands, the same delimiter token is inserted between successive context segments in the same causal order.
On the output side, a trainable value prediction head $W_{\text{val}} \in \mathbb{R}^{K\times d}$ (i.e., $2048 \times d$) replaces the full-vocabulary LM head for target positions.
For more stable training, both $E_{\text{val}}$ and $W_{\text{val}}$ are initialized from a consecutive block of pretrained Llama parameters.
This keeps the scale of the pretrained parameters, while the actual prediction space remains the local value vocabulary rather than the native full text vocabulary.

\subsubsection{Backbone Adaptation and Training}

The frozen backbone is Llama-3.2-3B~\cite{grattafiori2024llama3}, adapted via LoRA~\cite{hu2022lora} inserted into attention and MLP projections.
A single set of adapters is shared across all subband tasks and all four decomposition levels.
This shared-adapter design uses fewer trainable parameters, reduces training cost, and improves stability.

Given input embeddings $X$ with sequence length $T$, the backbone produces hidden states $h_{1:T}$.
At each target position $i \in \Omega$, the value prediction head computes logits $\ell_i = W_{\text{val}}\, h_i \in \mathbb{R}^K$, and the training loss is standard cross-entropy over the local value range:
\begin{equation}
\mathcal{L}_{\mathrm{CE}}=-\frac{1}{|\Omega|}\sum_{i\in\Omega}\log \mathrm{softmax}(\ell_i)_{z_i}.
\end{equation}
The estimated coding rate in bits per symbol is $\mathcal{L}_{\mathrm{CE}}/\ln 2$.
During actual coding, the arithmetic coder uses the same distributions in the same order.

Training and testing are both patch based.
After wavelet transformation, a patch from subband $b$ at level $l$ is denoted by
\begin{equation}
P_l^b \in \mathbb{Z}^{3\times h_p\times w_p}.
\end{equation}
Each training image is first wavelet transformed at full resolution to produce subbands, and patches are then cropped on the resulting wavelet-domain subbands.
Training samples the 13 subband tasks through an epoch-wise shuffled task queue:
\begin{equation}
\mathcal{Q}=\{(4,LL)\}\cup\{(l,HL),(l,LH),(l,HH)\}_{l=1}^{4}.
\end{equation}
For each sampled task in $\mathcal{Q}$, a random patch is drawn from the corresponding subband.
This random patch training still covers all 13 subband modeling tasks.

\begin{table}[t]
  \centering
  \caption{Implementation settings.}
  \label{tab:impl}
  \setlength{\tabcolsep}{3pt}
  \renewcommand{\arraystretch}{1.08}
  \footnotesize
  \begin{tabular}{@{}ll@{}}
    \toprule
    Item & Setting \\
    \midrule
    Backbone & Llama-3.2-3B (frozen) \\
    Adaptation & LoRA (shared across all tasks) \\
    LoRA rank / alpha / dropout & 128 / 256 / 0.05 \\
    Target modules & q/k/v/o + gate/up/down proj. \\
    Color transform & reversible YCoCg \\
    Wavelet transform & 4-level fixed 5/3 lifting \\
    Value range $K$ / offset $o$ & 2048 / 512 \\
    Value embedding init & consecutive pretrained block \\
    Patch size / batch / grad.\ accum. & 48 / 1 / 16 \\
    Optimizer / learning rate & AdamW / $10^{-5}$ \\
    Hardware & 8$\times$RTX\,3090 (bf16) \\
    Peak memory (single card) & 16.2\,GB (Kodak) / 20.9\,GB (DIV2K) \\
    \bottomrule
  \end{tabular}
\end{table}

\section{Experiments}

\subsection{Experimental Settings}

\subsubsection{Datasets}
Training uses three natural-image datasets: the NIC training set (607{,}715 images at $256\times 256$), the DIV2K training split (800 images at approximately 2K resolution)~\cite{agustsson2017div2k}, and the Flickr2K training split (2{,}650 images)~\cite{lim2017edsr}.
We use a two-stage schedule.
The model is first pretrained on the lower-resolution NIC corpus and then finetuned on the higher-resolution DIV2K and Flickr2K images.
All training images are disjoint from the held-out evaluation sets.
Model selection uses a separate validation split.

Evaluation is performed on three standard benchmarks: Kodak~\cite{kodak2013} (24 images, $768\times 512$), the professional track of CLIC~\cite{clic2020challenge} (41 images), and the DIV2K validation set~\cite{agustsson2017div2k} (100 images).
Images are converted to YCoCg and padded to multiples of 16 for four-level wavelet decomposition.
The rates reported for WaveLLM are measured from the written bitstreams and normalized by the original image dimensions before padding.

\subsubsection{Training Details}
Detailed implementation settings are summarized in Table~\ref{tab:impl}.
The backbone is Llama-3.2-3B with frozen pretrained weights, and LoRA is applied to the q\_proj, k\_proj, v\_proj, o\_proj, gate\_proj, up\_proj, and down\_proj modules. The rank, alpha, and dropout are 128, 256, and 0.05, respectively.
The local value range uses $K=2048$ with offset $o=512$.
The patch size in Table~\ref{tab:impl} refers to crops taken from wavelet-domain subbands after full-image transformation.
Training uses batch size 1 with gradient accumulation 16, AdamW, and a learning rate of $10^{-5}$ on 8 RTX 3090 GPUs in bf16.

\subsubsection{Baseline Codecs}
The baselines cover three groups.
Traditional codecs include PNG, JPEG2000, JPEG-LS, FLIF, and JPEG XL~\cite{png2003,taubman2002jpeg2000,jpegls1998,weinberger2000locoi,sneyers2016flif,jpegxl2019}.
Learned lossless codecs include L3C, RC, iWave, iVPF, DLPR, ArIB-BPS, and CALLIC~\cite{mentzer2019l3c,mentzer2020learning,ma2020iwave,zhang2021ivpf,bai2022dlpr,zhang2024bps,li2025callic}.
LLM-based baselines include Del\'etang et al., P$^2$-LLM, and the visual-prompt method~\cite{deletang2023lmcompression,chen2024p2llm,du2025visualprompt}.

\subsection{Main Results}

\subsubsection{Comparison with Existing Methods}
Table~\ref{tab:reference_results} presents a comparison between WaveLLM and traditional, learned, and LLM-based methods.
Table~\ref{tab:reference_results} reports WaveLLM rates alongside rates quoted from prior work. Table~\ref{tab:runtime} combines our iWave and WaveLLM measurements with runtime figures quoted from prior work; Tables~\ref{tab:domain_comparison} and~\ref{tab:fixed_codec} report measurements from our setup.

WaveLLM with a 3B backbone yields lower rates than both P$^2$-LLM and the visual-prompt method, even though both employ 8B backbones.
On Kodak, WaveLLM achieves 2.51~bpsp versus 2.83~bpsp for both LLM competitors, a reduction of 11.3\%.
On CLIC.pro and DIV2K, it reaches 2.24 and 2.27~bpsp versus 2.35/2.25 and 2.51/2.29~bpsp, respectively.
WaveLLM is also competitive with CALLIC~\cite{li2025callic}, a strong learned method that uses content-adaptive per-image fine-tuning.
Among strong non-LLM baselines, WaveLLM also improves over JPEG~XL, the strongest traditional codec in Table~\ref{tab:reference_results}, and the lossless iWave codec in the matched 5/3 setting, by 0.36 and 0.43~bpsp on Kodak.
Exact reconstruction was verified on all reported test sets, with no observed value-range overflow or out-of-range input or label indices.

\begin{table*}[t]
  \centering
  \small
  \caption{Lossless coding performance (bpsp) under reported settings; green percentages show bitrate increase over WaveLLM. Training data and auxiliary signals differ across papers.}
  \label{tab:reference_results}
  \setlength{\tabcolsep}{10.5pt}
  \renewcommand{\arraystretch}{1.12}
  \begin{tabular}{llccc}
    \toprule
    Category & Codec & Kodak & CLIC.pro & DIV2K \\
    \midrule
    \multirow{5}{*}{Traditional}
      & PNG & 4.35 \reldiff{73.3} & 3.93 \reldiff{75.4} & 4.23 \reldiff{86.3} \\
      & JPEG2000 & 3.19 \reldiff{27.1} & 2.93 \reldiff{30.8} & 3.12 \reldiff{37.4} \\
      & JPEG-LS & 3.16 \reldiff{25.9} & 2.82 \reldiff{25.9} & 2.99 \reldiff{31.7} \\
      & FLIF & 2.90 \reldiff{15.5} & 2.72 \reldiff{21.4} & 2.91 \reldiff{28.2} \\
      & JPEG~XL & 2.87 \reldiff{14.3} & 2.63 \reldiff{17.4} & 2.79 \reldiff{22.9} \\
    \midrule
    \multirow{7}{*}{Learned}
      & L3C~\cite{mentzer2019l3c} & 3.26 \reldiff{29.9} & 2.94 \reldiff{31.2} & 3.09 \reldiff{36.1} \\
      & RC~\cite{mentzer2020learning} & 3.38 \reldiff{34.7} & 2.93 \reldiff{30.8} & 3.08 \reldiff{35.7} \\
      & iWave~\cite{ma2020iwave} & 2.94 \reldiff{17.1} & 2.60 \reldiff{16.1} & 2.73 \reldiff{20.3} \\
      & iVPF~\cite{zhang2021ivpf} & --- & 2.54 \reldiff{13.4} & 2.68 \reldiff{18.1} \\
      & DLPR~\cite{bai2022dlpr} & 2.86 \reldiff{13.9} & 2.42 \reldiff{8.0} & 2.55 \reldiff{12.3} \\
      & ArIB-BPS~\cite{zhang2024bps} & 2.78 \reldiff{10.8} & 2.42 \reldiff{8.0} & 2.55 \reldiff{12.3} \\
      & CALLIC~\cite{li2025callic} & 2.54 \reldiff{1.2} & 2.30 \reldiff{2.7} & 2.46 \reldiff{8.4} \\
    \midrule
    \multirow{4}{*}{LLM-based}
      & Del\'etang et al.~\cite{deletang2023lmcompression} & 3.96 \reldiff{57.8} & 3.89 \reldiff{73.7} & 4.17 \reldiff{83.7} \\
      & P$^2$-LLM (8B)~\cite{chen2024p2llm} & 2.83 \reldiff{12.7} & 2.35 \reldiff{4.9} & 2.51 \reldiff{10.6} \\
      & Visual Prompt (8B)~\cite{du2025visualprompt} & 2.83 \reldiff{12.7} & 2.25 \reldiff{0.4} & 2.29 \reldiff{0.9} \\
      & \textbf{WaveLLM (3B, ours)} & \textbf{2.51} & \textbf{2.24} & \textbf{2.27} \\
    \bottomrule
  \end{tabular}
\end{table*}

\subsubsection{Computational Cost}
\label{sec:runtime}

Table~\ref{tab:runtime} reports runtime and computational cost on Kodak.
Because hardware, implementation details, and model sizes differ across papers, we interpret these numbers qualitatively.
WaveLLM requires about $2.6\times10^7$~kMACs/pixel and takes 65.1~s to encode and 152.8~s to decode each Kodak image in our implementation.
Although its 3B backbone is smaller than the 8B backbones used by prior LLM-based methods, end-to-end runtime remains hardware- and implementation-dependent.
WaveLLM parallelizes its 13 subband tasks across 8~RTX\,3090 GPUs, and its peak memory of about 16.2~GB per card allows the full pipeline to fit on 24~GB consumer GPUs.
This implementation also benefits from patch-wise parallel processing and an optimized arithmetic coder.
By contrast, the iWave baseline follows serial pixel-wise inference and is measured on a single RTX\,3090 in our environment.
The complexity and timing figures for Del\'etang et al.\ are quoted from~\cite{du2025visualprompt} rather than measured by us, as the original model weights are not publicly available.
These measurements describe the current implementation; runtime reduction remains future work.

\begin{table}[tb]
  \centering
  \small
  \caption{Per-image runtime on Kodak ($768\times512$), including arithmetic coding and I/O. iWave and WaveLLM are measured in our environment; other entries are quoted from prior work on different hardware. The kMACs/pixel column reports multiply-accumulate operations per output pixel where available.}
  \label{tab:runtime}
  \setlength{\tabcolsep}{3.5pt}
  \renewcommand{\arraystretch}{1.1}
  \begin{tabular}{lcccc}
    \toprule
    Method & kMACs/px & Enc.\,(s) & Dec.\,(s) & Hardware \\
    \midrule
    iWave (PixelCNN) & --- & 582.9 & 611.6 & RTX\,3090 \\
    Del\'etang et al.\ (8B) & $2.1\!\times\!10^7$ & 10.44 & 288.0 & 4$\times$A100 \\
    P$^2$-LLM (8B) & --- & 14.9 & 273.1 & 8$\times$A800 \\
    Visual Prompt (8B) & $4.2\!\times\!10^7$ & 21.1 & 495.6 & 4$\times$A100 \\
    \textbf{WaveLLM (3B)} & $2.6\!\times\!10^7$ & 65.1 & 152.8 & 8$\times$RTX\,3090 \\
    \bottomrule
  \end{tabular}
\end{table}

\subsection{Analysis of the Proposed Design}

\subsubsection{Why the Wavelet Domain}
\label{sec:domain_comparison}
To test directly whether the coding domain matters for LLM-based entropy modeling, we implement two alternative variants under the same general paradigm: a pixel-domain variant that performs next-token prediction on rasterized RGB values, and a residual-based variant that encodes the residual between the original image and a lossy reconstruction, using the global lossy image and the corresponding local reconstruction patch as visual prompts.
Table~\ref{tab:domain_comparison} summarizes results across all three benchmarks. Under the same Llama-3.2-3B backbone and next-token entropy-modeling framework, the wavelet-domain formulation is consistently the strongest.
This comparison is especially relevant because WaveLLM is, to the best of our knowledge, the first LLM-based lossless compressor formulated directly in the wavelet domain.

The pixel-domain variant is already competitive, indicating that the value interface and autoregressive LLM framework remain effective beyond the wavelet setting.
However, the wavelet-domain formulation outperforms it by 0.21, 0.11, and 0.17~bpsp on Kodak, CLIC.pro, and DIV2K, respectively.
The residual-based variant yields higher rates under our implementation, and its total rate includes both the lossy base bits and the coded residual bits.
We do not introduce an explicit parametric density model in the residual domain, which may limit how fully the model capacity is used in that setting.
Overall, the wavelet-domain formulation performs best among the tested variants.

\begin{table}[t]
  \centering
  \small
\caption{Coding-domain comparison under the same 3B setup. Pixel uses rasterized RGB; Residual codes residuals using the global lossy image and corresponding local reconstruction patch as prompts; Wavelet codes integer subband coefficients with same-level context. Green percentages indicate relative bitrate increase over Wavelet. Lower is better.}
  \label{tab:domain_comparison}
  \setlength{\tabcolsep}{5pt}
  \renewcommand{\arraystretch}{1.1}
  \begin{tabular}{@{}lp{26mm}ccc@{}}
    \toprule
    Domain & Side information & Kodak & CLIC.pro & DIV2K \\
    \midrule
    Pixel & none & 2.72 \reldiff{8.4} & 2.35 \reldiff{4.9} & 2.44 \reldiff{7.5} \\
    Residual & lossy thumb. & 2.97 \reldiff{18.3} & 2.43 \reldiff{8.5} & 2.48 \reldiff{9.3} \\
    Wavelet & same-level context & \textbf{2.51} & \textbf{2.24} & \textbf{2.27} \\
    \bottomrule
  \end{tabular}
\end{table}

\subsubsection{Wavelet-Domain Backbone-Scale Analysis}
\label{sec:fixed_codec}
\begin{figure}[tb]
  \centering
  \includegraphics[width=0.90\linewidth]{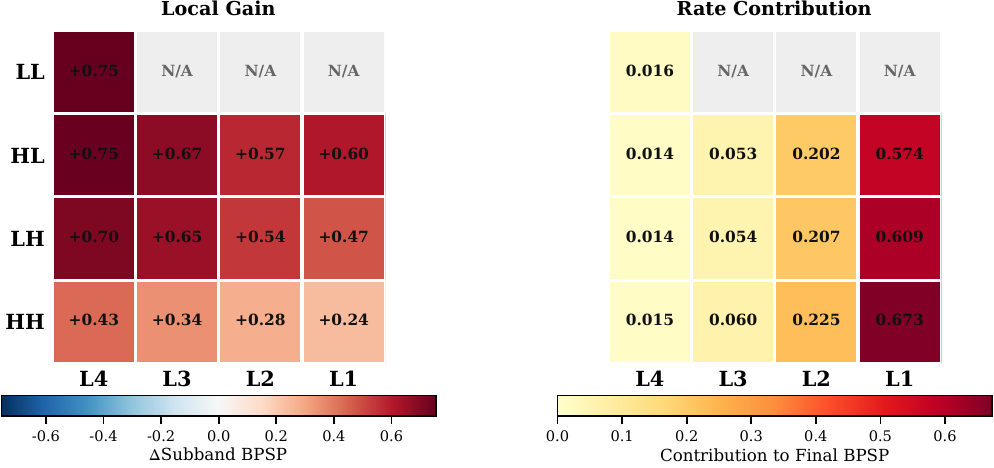}
  \caption{Wavelet-domain local gain and rate contribution on Kodak01. Rows follow LL/HL/LH/HH and columns follow Levels~4 to~1. Left: $\Delta$Subband~BPSP $=$ BPSP$_{\text{iWave}}$ $-$ BPSP$_{\text{WaveLLM}}$; right: each subband's absolute contribution to the final WaveLLM bitrate. Warmer colors indicate larger values.}
  \label{fig:analysis}
  \Description{Two side-by-side 4 by 4 wavelet-structured heatmaps for Kodak01, showing local coding gain on the left and each subband's absolute contribution to the final WaveLLM bitrate on the right.}
\end{figure}

Table~\ref{tab:fixed_codec} shows coding performance for entropy models within the same 5/3 wavelet codec.
The first row uses PixelCNN to autoregressively model the coefficients, while the other rows use LLM backbones of different sizes.
Compared with PixelCNN, WaveLLM-3B reduces the rate from 2.94 to 2.51~bpsp on Kodak ($-$0.43, $-$14.6\%), from 2.60 to 2.24~bpsp on CLIC.pro ($-$0.36, $-$13.8\%), and from 2.73 to 2.27~bpsp on DIV2K ($-$0.46, $-$16.8\%).
WaveLLM-1B already outperforms the PixelCNN baseline, showing that the wavelet-domain formulation remains effective at a smaller backbone scale.
Scaling from 1B to 3B yields further gains of 0.37/0.17/0.23~bpsp, indicating that model capacity still matters within this formulation.
The larger backbone provides more capacity for the shared adapters to specialize across all 13 subband types.
Figure~\ref{fig:analysis} shows each subband's local gain and contribution to the Kodak01 bitrate.
At every subband, the LLM yields a positive local gain over iWave, which indicates more accurate probability prediction throughout the hierarchy.
Low-frequency and coarse high-frequency bands benefit most, whereas shallow Level-1 bands dominate the total rate because they contain most coded symbols.

\begin{table}[tb]
  \centering
  \small
  \caption{Lossless coding performance (bpsp) within the same reversible wavelet codec. The transform, coding order, and arithmetic coder are fixed; only the entropy model changes. Green percentages indicate the relative bitrate increase over WaveLLM-3B. Lower is better.}
  \label{tab:fixed_codec}
  \setlength{\tabcolsep}{4pt}
  \renewcommand{\arraystretch}{1.1}
  \begin{tabular}{lccc}
    \toprule
    Entropy model / Backbone & Kodak & CLIC.pro & DIV2K \\
    \midrule
    PixelCNN (iWave) & 2.94 \reldiff{17.1} & 2.60 \reldiff{16.1} & 2.73 \reldiff{20.3} \\
    WaveLLM-1B & 2.88 \reldiff{14.7} & 2.41 \reldiff{7.6} & 2.50 \reldiff{10.1} \\
    \textbf{WaveLLM-3B} & \textbf{2.51} & \textbf{2.24} & \textbf{2.27} \\
    \bottomrule
  \end{tabular}
\end{table}

\subsubsection{Task-Specific Distribution Analysis}
\label{sec:analysis}

\begin{figure}[tb]
  \centering
  \includegraphics[width=0.84\linewidth]{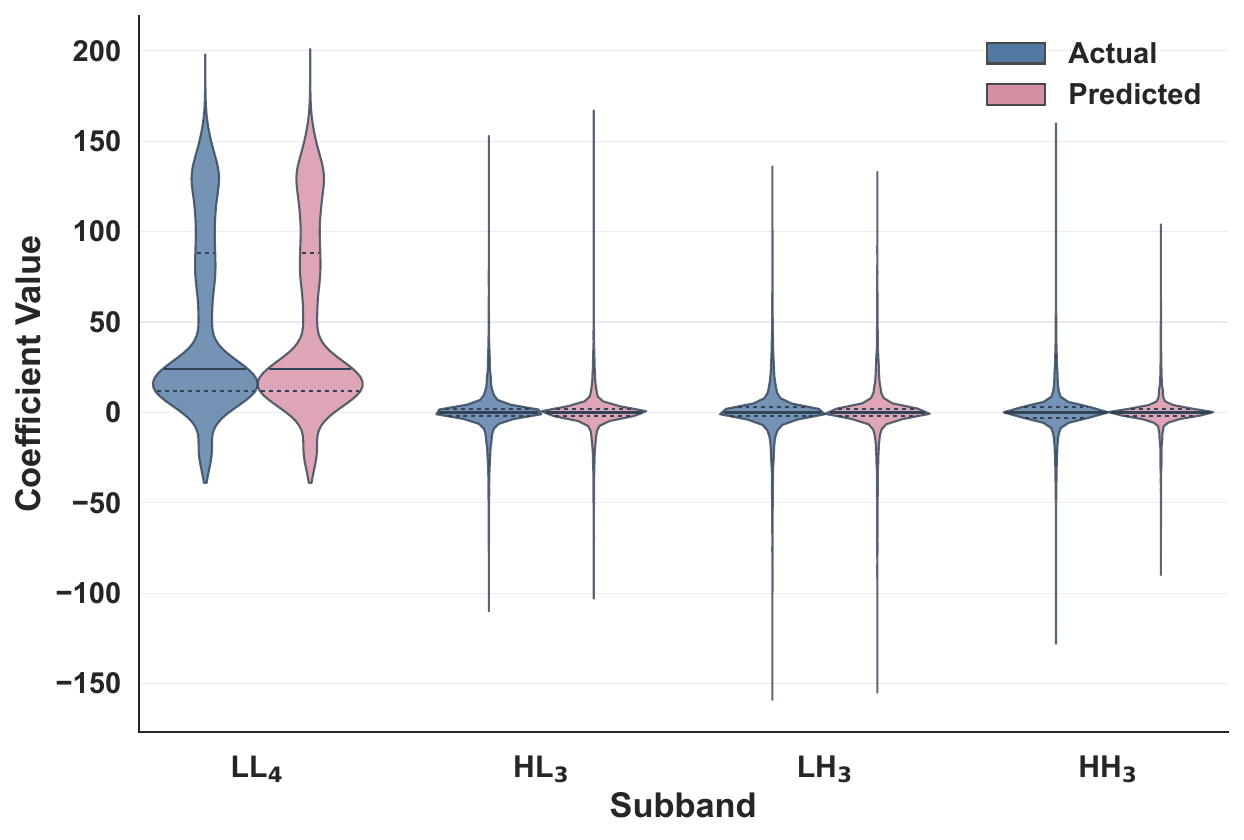}
  \caption{Actual coefficient and LLM-predicted peak-value distributions for four Kodak01 subbands. Each pair compares the two distributions at matched positions. $LL_4$ is shown in coefficient space after converting MED residuals back; the remaining pairs show $HL_3$, $LH_3$, and $HH_3$.}
  \label{fig:vio}
  \Description{A violin plot comparing actual coefficient-value distributions and LLM-predicted peak-value distributions for Kodak01 on four subbands: LL4 on the left, followed by HL3, LH3, and HH3. Each subband has two violins labeled Actual and Predicted.}
\end{figure}

Figure~\ref{fig:vio} compares actual and LLM-predicted peak-value distributions for four subbands: $LL_4$, $HL_3$, $LH_3$, and $HH_3$.
Predicted peaks are task-specific and align with actual distributions, indicating that one shared LoRA-adapted backbone learns subband-specific statistics.
For $LL_4$, the plotted distributions are shown in coefficient space after converting the MED-based coding results back from residual representation.
The $LL_4$ coefficients follow a much broader distribution, while $HL_3$, $LH_3$, and $HH_3$ remain sharply peaked around zero.
Predictions for the three high-frequency bands differ in tails and spread despite similar overall shapes.
Task prompts and decoded context steer the shared backbone toward subband-specific distributions rather than a single generic prior across the 13 entropy-modeling tasks.
This lets one shared backbone model all 13 subband tasks.

\subsection{Ablation Study}
\label{sec:ablation}

Table~\ref{tab:ablation} isolates four design choices on Kodak and CLIC.pro; each row changes one component relative to the full system~(a).
The ablations address four themes: task identity, the numerical interface, context design, and low-frequency residual modeling.

\subsubsection{Task Identity and Prompt}
Removing text prompts causes the largest single-component degradation (row~b, $+$0.11/$+$0.21~bpsp on Kodak/CLIC.pro).
Without task-specific instructions, the shared backbone lacks an explicit task-identity signal and tends to approximate an average distribution across the 13 subband tasks.
These results support the effectiveness of the proposed task prompt.

\subsubsection{Numerical Interface}
Using the native Llama vocabulary in place of the value interface increases the rate by $+$0.07/$+$0.08~bpsp (row~c).
Neighboring coefficients may map to distant token IDs, and language-trained embeddings introduce a domain gap. The native 128K vocabulary also defines a less compact probability space.
The proposed local value interface avoids these issues and yields more stable coefficient modeling.

\subsubsection{Context Design}
We test two context variants: same-level $LL_l$ only (row~d, $+$0.03/$+$0.06~bpsp on Kodak/CLIC.pro) and same-level plus parent HF bands (row~e, $+$0.23/$+$0.27~bpsp).
Using only $LL_l$ causes a small loss because it provides most of the useful low-frequency context.
Adding parent-level HF bands is more harmful, likely because the longer sequence weakens local attention and the coarser scale is less aligned with the target statistics.
These results support the current design based on same-level decoded context.

\subsubsection{Low-Frequency MED Residuals}
Omitting MED residual coding for $LL_4$ (row~f, $+$0.01/$+$0.03~bpsp on Kodak/CLIC.pro) causes a small but consistent loss.
Figure~\ref{fig:analysis} shows that $LL_4$ itself contributes only about 0.016~bpsp to the Kodak01 total rate.
Even so, MED makes the $LL_4$ representation more concentrated and closer to the high-frequency subbands, which appears to benefit end-to-end training.

\begin{table}[tb]
  \centering
  \small
  \caption{Ablations of WaveLLM (bpsp). Each row changes one design choice relative to the full system~(a). $\Delta$: rate increase (positive\,=\,worse). Val.\,IF: value interface. HF scope: same\,=\,same-level, LL\,=\,LL-only, +par.\,=\,same + parent-level.}
  \label{tab:ablation}
  \setlength{\tabcolsep}{2.2pt}
  \renewcommand{\arraystretch}{1.15}
  \footnotesize
  \begin{tabular}{@{}lcccccc@{}}
    \toprule
    & Prompt & Val.\,IF & HF scope & MED & Kodak\;($\Delta$) & CLIC.pro\;($\Delta$) \\
    \midrule
    (a) Full system & $\checkmark$ & ded. & same & $\checkmark$ & \textbf{2.51} & \textbf{2.24} \\
    \addlinespace[3pt]
    (b) $-$\,prompt & $\times$ & ded. & same & $\checkmark$ & 2.62\;($+$0.11) & 2.45\;($+$0.21) \\
    (c) $-$\,val.\,IF & $\checkmark$ & res. & same & $\checkmark$ & 2.58\;($+$0.07) & 2.32\;($+$0.08) \\
    (d) $-$\,HF ctx & $\checkmark$ & ded. & LL & $\checkmark$ & 2.54\;($+$0.03) & 2.30\;($+$0.06) \\
    (e) $+$\,parent HF & $\checkmark$ & ded. & same+par. & $\checkmark$ & 2.74\;($+$0.23) & 2.51\;($+$0.27) \\
    (f) $-$\,MED & $\checkmark$ & ded. & same & $\times$ & 2.52\;($+$0.01) & 2.27\;($+$0.03) \\
    \bottomrule
  \end{tabular}
\end{table}

\FloatBarrier

\section{Conclusion}

We presented WaveLLM, a wavelet-domain LLM entropy modeling framework for lossless image compression. It combines a coding-oriented numerical interface, task prompts, and same-level decoded context within a LoRA-adapted Llama-3.2-3B backbone. WaveLLM achieves 2.51, 2.24, and 2.27 bpsp on Kodak, CLIC.pro, and DIV2K, respectively, outperforming existing LLM-based compressors on all three benchmarks while remaining competitive with strong lossless codecs. These results suggest that wavelet subbands provide a more suitable representation for LLM-based probability modeling than the tested flat pixel and residual sequences. By separating image content into wavelet subbands and guiding each prediction with task prompts and decoded context, WaveLLM enables more accurate entropy estimation. Future work will consider distillation, quantization, and near-lossless coding.

\clearpage

\begin{acks}
This work was in part supported by the National Natural Science Foundation of China
under grants 62472399 and Open Fund of APKL of BIIP, IAI, Hefei Comprehensive
National Science Center under grants 24YGXT003.
\end{acks}

\bibliographystyle{ACM-Reference-Format}
\bibliography{mm2026_refs}

\end{document}